\documentclass[runningheads]{llncs}
\usepackage[pdftex]{graphicx}

\usepackage{tikz}
\usepackage{comment}
\usepackage{amsmath,amssymb} 
\usepackage{color}
\usepackage{mathtools}
\usepackage{subfigure}
\usepackage{algpseudocode}
\usepackage{algorithm}
\usepackage{wrapfig}
\usepackage{todonotes}
\usepackage[square,numbers]{natbib}
\usepackage[pagebackref,breaklinks,colorlinks,bookmarks=false]{hyperref}
\usepackage{overpic}
\usepackage{orcidlink}

\usepackage{multirow}
\usepackage{tabularx}
\usepackage{url}
\usepackage{booktabs}
\usepackage{gensymb}
\usepackage{adjustbox}

\newcolumntype{Y}{>{\centering\arraybackslash}X}
\begin{document}

\pagestyle{headings}
\mainmatter
\def\ECCVSubNumber{074}  

\title{AvatarPoser: Articulated Full-Body Pose Tracking from Sparse Motion Sensing} 

\titlerunning{AvatarPoser}
%
\author{Jiaxi Jiang\inst{1}\orcidlink{0000-0001-6000-4474},
Paul Streli\inst{1}\orcidlink{0000-0002-3334-7727},
Huajian Qiu\inst{1}\orcidlink{0000-0001-7792-0241},
Andreas Fender\inst{1}\orcidlink{0000-0002-5903-0736
},\\
Larissa Laich\inst{2}\orcidlink{0000-0001-7823-9730},
Patrick Snape\inst{2}\orcidlink{0000-0001-8844-3225},
Christian Holz\inst{1}\orcidlink{0000-0001-9655-9519}
}
\authorrunning{Jiang et al.}
%
\institute{Department of Computer Science, ETH Zurich, Switzerland 
\and
Reality Labs at Meta, Switzerland
\\
\url{https://github.com/eth-siplab/AvatarPoser}
}

\maketitle
\newcommand{\larissa}[1]{\todo[inline, bordercolor=blue,backgroundcolor=white]{[larissa]: {#1}}}
\newcommand{\jiaxi}[1]{\todo[inline, bordercolor=red,backgroundcolor=white]{[jiaxi]: {#1}}}
\newcommand{\ch}[1]{\todo[inline, bordercolor=orange,backgroundcolor=white]{[ch]: {#1}}}
\newcommand{\patrick}[1]{\todo[inline, bordercolor=yellow,backgroundcolor=white]{[patrick]: {#1}}}

\newcommand{\method}{\emph{AvatarPoser }}
\newcommand{\methodnospace}{\emph{AvatarPoser}}

\begin{abstract}
Today's Mixed Reality head-mounted displays track the user's head pose in world space as well as the user's hands for interaction in both Augmented Reality and Virtual Reality scenarios.
While this is adequate to support user input, it unfortunately limits users' virtual representations to just their upper bodies. Current systems thus resort to floating avatars, whose limitation is particularly evident in collaborative settings. To estimate full-body poses from the sparse input sources, prior work has incorporated additional trackers and sensors at the pelvis or lower body, which increases setup complexity and limits practical application in mobile settings. In this paper, we present \methodnospace, the first learning-based method that predicts full-body poses in world coordinates using only motion input from the user's head and hands. Our method builds on a Transformer encoder to extract deep features from the input signals and  decouples global motion from the learned local joint orientations to guide pose estimation.
To obtain accurate full-body motions that resemble motion capture animations, we refine the arm joints' positions using an optimization routine with inverse kinematics to match the original tracking input.
In our evaluation, \method achieved new state-of-the-art results in evaluations on large motion capture datasets (AMASS). At the same time, our method's inference speed supports real-time operation, providing a practical interface to support holistic avatar control and representation for Metaverse applications.

\end{abstract}

\section{Introduction}
\begin{figure}[ht!]
    \centering
    \includegraphics{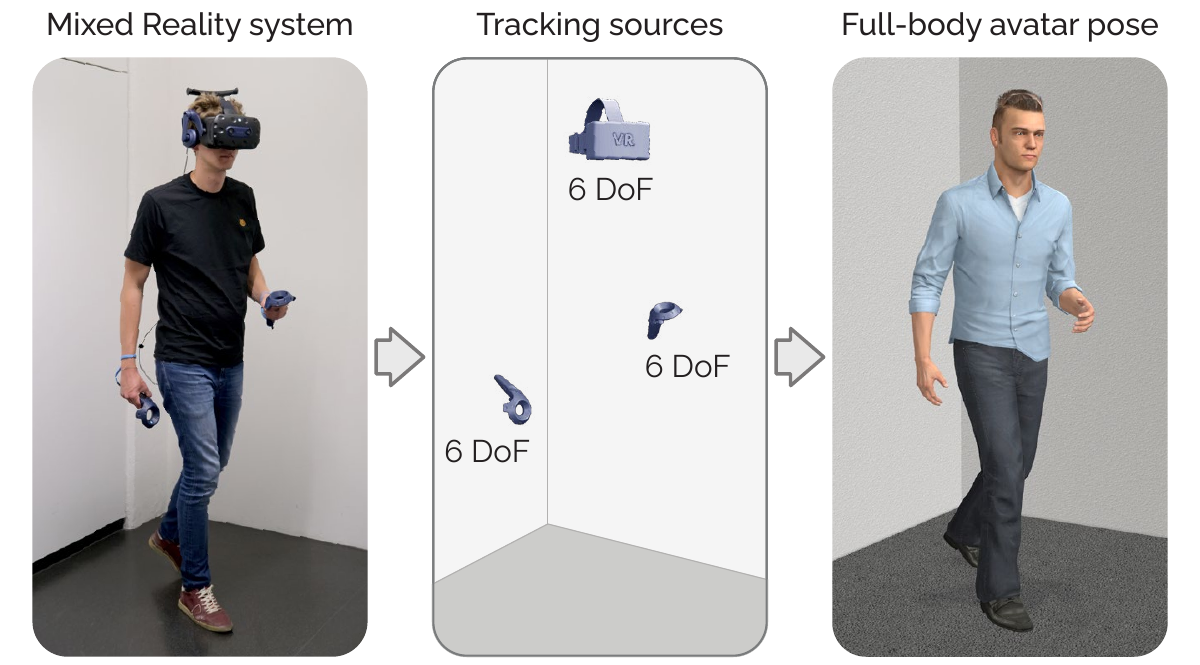}
    \caption{%
         We address the new problem of full-body avatar pose estimation from sparse tracking sources, which can significantly enhance embodiment, presence, and immersion in Mixed Reality.
         Our novel Transformer-based method \method takes as input only the positions and orientations of one headset and two handheld controllers (or hands), and generates a full-body avatar pose over 22 joints.
         Our method reaches state-of-the-art pose accuracy, while providing a practical interface into the Metaverse.
    }
    \label{fig:teaser}
\end{figure}

Interaction in today's Mixed Reality (MR) environments is driven by the user's head pose and input from the hands.
Cameras embedded in head-mounted displays (HMD) track the user's position inside the world and estimate articulated hand poses during interaction, which finds frequent application in Augmented Reality (AR) scenarios.
Virtual Reality (VR) systems commonly equip the user with two hand-held controllers for spatial input to render haptic feedback.
In both cases, even this sparse amount of tracking information suffices for interacting with a large variety of immersive first-person experiences.

However, the lack of complete body tracking can break immersion and reduce the fidelity of the overall experience as soon as interactions exceed manual first-person tasks.
This not just becomes evident as users see their own bodies during interaction in VR, but also in collaborative tasks in AR that necessarily limit the representation of other participants to their upper bodies, rendered to hover through space.
Studies on avatar appearances have shown the importance of holistic avatar representations to achieve embodiment~\cite{Waltemate2018} and to establish presence in the virtual environment~\cite{Heidicker2017}.
Applications such as telepresence or productivity meetings would greatly benefit from more holistic avatar representations that approach the fidelity of motion-capture animations.

This challenge will likely not be addressed by future hardware improvements, as MR systems increasingly optimize for mobile use outside controlled spaces that could accommodate comprehensive tracking.
Therefore, we cannot expect future systems to expand much on the tracking information that is available today.
While the headset's cameras may partially capture the user's feet in opportune moments with a wide field of view, head-mounted cameras are generally in a challenging location for capturing ego-centric poses~\cite{wang2021estimating}.

Animating a complex full-body avatar based on the sparse input available on today's platforms is a vastly underdetermined problem.
To estimate the complete set of joint positions from the limited tracking sources, previous work has constrained the extend of motion diversity~\cite{ahuja2021coolmoves} or used additional trackers on the user's body, such as a 6D pelvis tracker~\cite{yang2021lobstr} or several body-worn inertial sensors~\cite{DIP:SIGGRAPHAsia:2018}.
Dittadi et al.'s recent method estimates full-body poses from only the head and hand poses with promising results~\cite{dittadi2021full}.
However, since the method encodes all joints relative to the pelvis, it implicitly assumes knowledge of a fourth 3D input (i.e., the pelvis).

For practical application, existing methods for full-body avatar tracking come with three limitations:
(1) Most general-purpose applications use Inverse Kinematics (IK) to estimate full-body poses.
This often generates human motion that appears static and unnatural, especially for those joints that are far away from the known joint locations in the kinematic chain.
(2) Despite the goal of using input from only the head and hands, existing deep learning-based methods implicitly assume knowledge of the pelvis pose.
However, pelvis tracking may never be available in most portable MR systems, which increases the difficulty of full-body estimation.
(3) Even with a tracked pelvis joint, animations from estimated lower-body joints sometimes contain jitter and sliding artifacts.
These tend to arise from unintended movement of the pelvis tracker, which is attached to the abdomen and thus moves differently from the actual pelvis joint.

In this paper, we propose a novel Transformer-based method for full human pose estimation with only the sparse tracking information from the head and hand (or controller) poses as input.
With \methodnospace, we decouple the global motion from learned pose features and use it to guide our pose estimation.
This provides robust results in the absence of other inputs, such as pelvis location or inertial trackers.
To the best of our knowledge, our method is the first to recover the full-body motion from only the three inputs across a wide variety of motion classes.
Because the predicted end effector poses of an avatar accumulate errors through the kinematic chain, we optimize our initial parameter estimations through inverse kinematics.
This combination of our learning-based method with traditional model-based optimization strikes a good balance between full-body style realism and accurate hand control.

We demonstrate the effectiveness of \method on the challenging AMASS dataset.
Our proposed method achieves state-of-the-art accuracy on full-body avatar estimation from sparse inputs.
For inference, our network reaches rates of up to 662\,fps.
In addition, we test our method on data we recorded with an HTC VIVE system and find good generalization of \method to unseen user input.
Taken together, our method provides a suitable solution for practical applications that operate based on the available tracking information on current MR headsets for application in both, Augmented Reality scenarios and Virtual Reality environments.

\section{Related Work}

\noindent\textbf{Full-Body Pose Estimation from Sparse Inputs.}
Much prior work on full-body pose estimation from sparse inputs has used up to 6 body-worn inertial sensors~\cite{von2017sparse, DIP:SIGGRAPHAsia:2018, yi2021transpose, yi2022physical}.
Because these 6 IMUs are distributed over head, arms, pelvis and legs, motion capture becomes inflexible and unwieldly.
CoolMoves~\cite{ahuja2021coolmoves} was first to use input from only the headset and hand-held controllers to estimate full-body poses.
However, the proposed KNN-based method interpolates poses from a smaller dataset with only specific motion activities and it is unclear how well it scales to large datasets with diverse subjects and activities, also for inference.
LoBSTr~\cite{yang2021lobstr} used a GRU network to predict the lower-body pose from the past sequence of tracking signals of the head, hands, and pelvis, while it computes the upper-body pose to match the tracked end-effector transformations via an IK solver.
The authors also highlight the difficulty of developing a system for estimations from 3 sources only, especially when distinguishing a wide range of human poses due to the large amount of ambiguity.
More recently, Dittadi et al. proposed a VAE-based method to generate plausible and diverse body poses from sparse input~\cite{dittadi2021full}.
However, their method implicitly uses knowledge of the pelvis as a fourth input location by encoding all joints relative to the pelvis, which leaves the highly ill-posed problem with only three inputs unsolved.\mbox{}\\

\noindent\textbf{Vision Transformer.} Transformers have achieved great success in their initial application in natural language processing~\cite{vaswani2017attention, devlin2019bert, dai2019transformer}.
The use of Transformer-based models has also significantly
improved the performance on various computer vision tasks such as image classification~\cite{dosovitskiy2021an,liu2021swin,fan2021multiscale}, image restoration~\cite{liang2021swinir, zamir2022restormer, wang2022uformer}, object detection~\cite{carion2020end, zhu2020deformable, sun2021rethinking}, and object tracking~\cite{meinhardt2022trackformer, zhao2022tracking, sun2020transtrack}.
In the area of human pose estimation, METRO~\cite{lin2021end} was first to apply Transformer models to vertex-vertex and vertex-joint interactions for 3D human pose and mesh reconstruction from a single image.
PoseFormer~\cite{zheng20213d} and ST-Transformer~\cite{aksan2020motiontransformer} used Transformers to capture both body joint correlations and temporal dependencies.
MHFormer~\cite{li2022mhformer} leveraged the spatio-temporal representations of multiple pose hypotheses to predict 3D human pose from monocular videos.
In contrast to their offline setting where the complete time series of motions are available, our method focuses on the practical scenario where streaming data is processed by our Transformer in real-time without looking ahead.\mbox{}\\

\noindent\textbf{Inverse Kinematics.}
Inverse kinematics (IK) is the process of calculating the variable joint parameters to produce a desired end-effector location.
IK has been extensively studied in the past, with various applications in robotics~\cite{goldenberg1985complete, parker1989inverse, wang1991combined, ruppel2018cost, maric2021riemannian} and computer animation~\cite{zhao1994inverse, grochow2004style, sumner2005mesh, parger2018human,finalik}.
Because no analytical solution usually exists for an IK problem, the most common way to solve the problem is through numerical methods via iterative optimization, which is costly.
To speed up computation, several heuristic methods have been proposed to approximate the solution~\cite{aristidou2011fabrik, luenberger1984linear, rokbani2015ik, ccavdar2013new}.
Recently, learning-based IK solutions have attracted attention~\cite{csiszar2017solving, bocsi2011learning, villegas2018neural, ren2020learning, ames2022ikflow, duka2014neural}, because they can speed up inference.
However, these methods are usually restricted to a scenario with a known data distribution and may not generalize well.
To overcome this problem, recent works have combined IK with deep learning to make the prediction more robust and flexible~\cite{starke2019neural, li2020mobile, yang2021lobstr, kang2021rcik, li2021hybrik, zhang2022couch}.
Our proposed method combines a deep neural network with IK optimization, where the IK component of our method refines the arm articulation to match the tracked hand positions from the original input (i.e., position of the hands or hand-held controllers).\mbox{}\\

\section{Method}

\subsection{Problem Formulation}

Although MR systems differ in the tracking technology they rely on, the global positions in Cartesian coordinates $\mathbf{p}^{1\times3}$ and orientations in axis-angle representation $\mathbf{\Phi}^{1\times3}$ of the headset and the hand-held controllers or hands are generally available.
From these, \method reconstructs the position of the articulated joints of the user's full body within the world $\mathbf{w}$.
This mapping $f$ is described through the following equation,
\begin{equation}
    \mathbf{T}_{1:\tau}^{1:F} = f(\{\mathbf{p^w},\mathbf{\Phi^w}\}_{1:\tau}^{1:S}),
\end{equation}
where $S$ corresponds to the number of joints tracked by the MR system, $F$ is the number of joints of the full-body skeleton, $\tau$ matches the number of observed MR frames that are considered from the past, and $\mathbf{T}\in SE(3)$ is the body joint pose which is represented by $\mathbf{T} = \{\mathbf{p},\mathbf{ \Phi}\}$. 

Specifically, we use the SMPL model~\cite{loper2015smpl} to represent and animate our human body pose.
We use the first 22 joints defined in the kinematic tree of the SMPL human skeleton and ignore the pose of fingers similar to previous work~\cite{dittadi2021full}.

\subsection{Input and Output Representation}

Since the 6D representation of rotations has proved effective for training neural networks due to its continuity~\cite{zhou2019continuity}, we convert the default axis-angle representation $\mathbf{\Phi}^{1\times3}$ in the SMPL model to the rotation matrix $\mathbf{R}^{3\times3}$ and discard the last row to get the 6D rotation representation $\mathbf{\theta}^{1\times6}$. During development, we observed that this 6D representation produces smooth and robust rotation predictions.

In addition to the accessible positions $\mathbf{p}^{1\times3}$ and orientations $\mathbf{R}^{3\times3}$ of the headset and hands, we also calculate the corresponding linear and angular velocities to obtain a signal of temporal smoothness.
The linear velocity $\mathbf{v}$ is given by backward finite difference at each time step $t$:
\begin{equation}
    \mathbf{v}_t = \mathbf{p}_{t} - \mathbf{p}_{t-1}
\end{equation}
Similar, the angular velocity $\mathbf{\Omega}$ can be calculated by:
\begin{equation}
\mathbf{\Omega}_t = \mathbf{R}^{-1}_{t-1}\mathbf{R}_t
\end{equation}
followed by also converting to its 6D representation $\mathbf{\omega}^{1\times6}$. As a result, the final input representation is a concatenated vector of position, linear velocity, rotation, and angular velocity from all given sparse inputs, which we write as:

\begin{equation}
    \textbf{X}_{t}^{1\times 18S}=[\{\mathbf{p}_{t}^1,\mathbf{v}_{t}^1,\mathbf{\theta}_{t}^1,\mathbf{\omega}_{t}^1\}^{1\times18}, \dots, \{\mathbf{p}_{t}^S,\mathbf{v}_{t}^S,\mathbf{\theta}_{t}^S,\mathbf{\omega}_{t}^S\}^{1\times18}]
\end{equation}
Therefore, when the number of sparse trackers $S$ equals $3$, the number of input features at each time step is 54.

The output of our rotation-based pose estimation network is the local rotation at each joint with respect to the parent joints $\mathbf{\theta}_\text{local}$.
The rotation value at the pelvis, which is the root of the SMPL model, refers to the global orientation $\mathbf{\theta}_\text{global}$.
As we use 22 joints to represent the full-body motion, the output dimension at each time step is 132.

\subsection{Overall Framework for Avatar Full-Body Pose Estimation}

\begin{figure}[t]
    \centering
\includegraphics{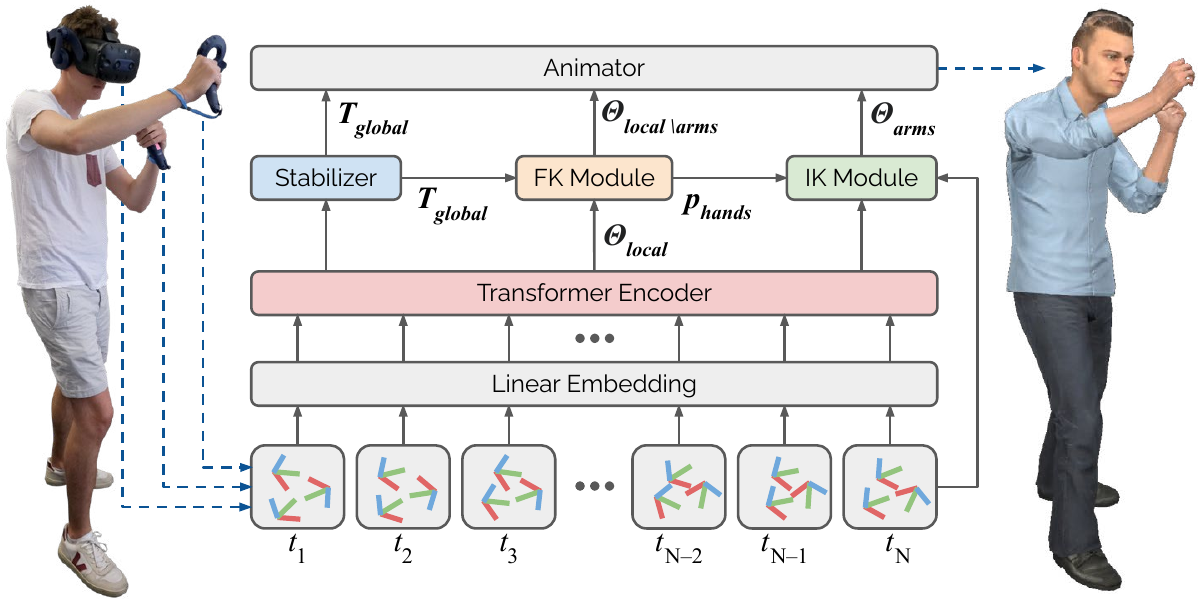}

    \caption{The framework of our proposed \method for Mixed Reality avatar full-pose estimation integrates four parts: a Transformer Encoder, a Stabilizer, a Forward-Kinematics Module, and an Inverse-Kinematics Module.
    The Transformer Encoder extracts deep pose features from previous time step signals from the headset and hands, which are split into global and local branches and correspond to global and local pose estimation, respectively.
    The Stabilizer is responsible for global motion navigation by decoupling global orientation from pose features and estimating global translation from the head position through the body's kinematic chain.
    The Forward-Kinematics Module calculates joint positions from a human skeleton model and a predicted body pose.
    The Inverse-Kinematics Module adjusts the estimated rotation angles of joints on the shoulder and elbow to reduce hand position errors.}
    \label{fig:architecture}
\end{figure}

Fig.~\ref{fig:architecture} illustrates the overall framework of our proposed method \methodnospace.
\method is a time series network that takes as input the 6D signals from the sparse trackers over the previous $N-1$ frames and the current $N^\text{th}$ frame and predicts global orientation of the human body as well as the local rotations at each joint with respect to its parent joint.
Specifically, \method consists of four components: a Transformer Encoder, a Stabilizer, a Forward-Kinematics (FK) Module, and an Forward-Kinematics (IK) Module.
We designed the network such that each component solves a specific task. \mbox{}\\

\noindent\textbf{\small Transformer Encoder.}
Our method builds on a Transformer model to extract the useful information from time-series data, following its benefits in efficiency, scalability, and long-term modeling capabilities.
We particularly leverage the Transformer's self-attention mechanism to distinctly capture global long-range dependencies in the data.
Specifically, given the input signals, we apply a linear embedding to enrich the features to 256 dimensions.
Next, our Transformer Encoder extracts deep pose features from previous time steps from the headset and hands, which are shared by the Stabilizer for global motion prediction, and a 2-layer multi-layer perceptron (MLP) for local pose estimation, respectively. We set the number of heads to 8 and the number of self-attention layers to 3.\mbox{}\\

\noindent\textbf{\small Stabilizer.}
The Stabilizer is a 2-layer MLP that takes as input the 256-dimensional pose features from our Transformer Encoder.
We set the number of nodes in the hidden layer to 256.
The output of the network produces the estimated global orientation represented as the rotation of the pelvis; therefore, it is responsible for global motion navigation by decoupling global orientation from pose features and obtaining global translation from the head position through the body kinematic chain.
Although it may be intuitive and possible to calculate the global orientation from a given head pose through the kinematic chain, the user's head rotation is often independent of the motions of other joints.
As a result, the global orientation at the pelvis is sensitive to the rotation of the head.
Considering the scenario where a user stands still and only rotates their head, it is likely that the global orientation may have a large error, which often results in a floating avatar.\mbox{}\\

\noindent\textbf{\small Forward-Kinematics Module.} The Forward-Kinematics (FK) Module calculates all joint positions given a human skeleton model and predicted local rotations as input.
While rotation-based methods provide robust results without the need to reproject onto skeleton constraints to avoid bone stretching and invalid configurations, they are prone to error accumulating along the kinematic chain. Training the network without FK could only minimize the rotation angles, but would not consider the actually resulting joint positions during optimization.\mbox{}\\

\begin{figure}[t]
    \centering
    \includegraphics{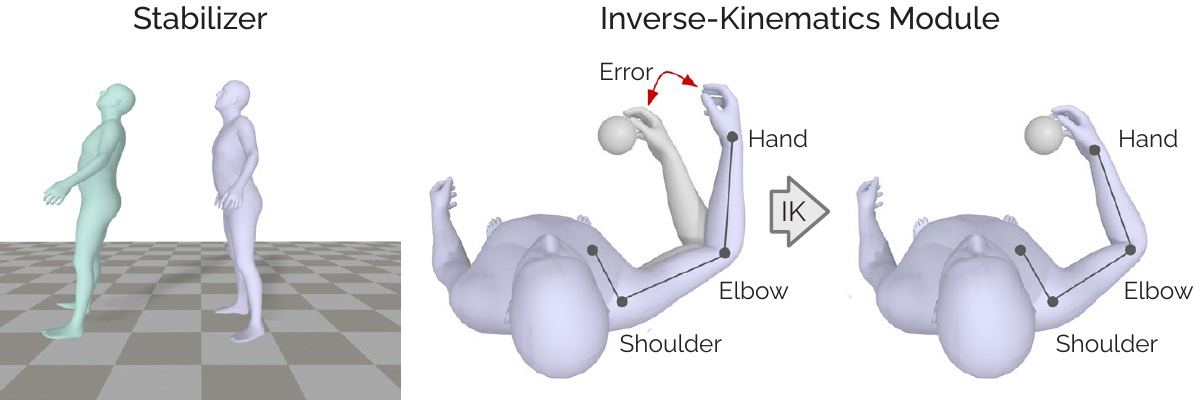}

    \caption{%
        \textbf{Left:} Our Stabilizer predicts global orientation and, thus, global motion (right avatar) that is significantly more robust than simply aligning the head of predicted body with the known input orientation (left avatar).
        \textbf{Right:} To account for accumulated errors along the joint hierarchy, we integrate an additional IK step to optimize the end-effectors' locations and match their target positions.
        }
    \label{fig:ik}
\end{figure}

\noindent\textbf{\small Inverse-Kinematics Module.} A main problem of rotation-based pose estimation is that the prediction of end-effectors may deviate from their actual location---even if the end effector served as a known input, such as in the case of hands.
This is because for end-effectors, the error accumulates along the kinematic chain.
Accurately estimating the position of end-effectors is particularly important in MR, however, because hands typically often used for providing input and even small errors in position can significantly disturb interaction with virtual interface elements.
To account for this, we integrate a separate IK algorithm that adjusts the arm limb positions according to the known hand positions.

Our methods performs IK-based optimization based on the estimated parameters output by our neural network.
This combines the individual benefits of both approaches as explored in prior work (e.g.,~\cite{li2020mobile,yang2021lobstr}).
Specifically, after our network produces an output, our IK Module adjusts the estimated rotation angles of joints on the shoulder and elbow to reduce the error of hand positions as shown in Fig.~\ref{fig:ik}.
We thereby fix the position of the shoulder and do not optimize the other rotation angles, because we found the resulting overall body posture to appear more accurate than the output of the IK algorithm.

Given the initial rotation values $ \mathbf{\theta}_{0} = \{\mathbf{\theta}_{0}^{\text{shoulder}}, \mathbf{\theta}_{0}^{\text{elbow}}\}$ estimated from our Transformer network, we calculate the positional error $\mathbf{E}$ of the hand according to the input signals and estimated hand position through the FK Module by
\begin{equation}
    \mathbf{E}(\theta_i) = \left\lVert \mathbf{p}^{\text{hand}}_{\text{gt}} - \text{FK}(\theta_i)^{\text{hand}} \right\rVert^2_2
\end{equation}
Then the rotation value is updated by:
\begin{equation}
    \mathbf{\theta}_{i+1} = \mathbf{\theta}_{i} + \eta\cdot f(\nabla\mathbf{E}(\theta_i))
\end{equation}
where $\eta$ is the learning rate and $f(\cdot)$ is decided by the specific optimizer.
To enable fast inference for real application, we stop the optimization after a fixed number of iterations.

There are several classical non-linear optimization algorithms that are suitable for optimizing inverse kinematics problems, such as Gauss-Newton method or the Levenberg-Marquardt method~\cite{more1978levenberg}.
In our experiment, we leverage the Adam optimizer~\cite{kingma2015adam} due to its compatibility with Pytorch.
We set the learning rate as $1\times10^{-3}$. \mbox{}\\

\noindent\textbf{\small Loss Function.} The final loss function is composed of an L1 local rotational loss, an L1 global orientation loss, and an L1 positional loss, denoted by:
\begin{equation}
    \mathbf{L}_{total} = \mathbf{\lambda}_{ori} \mathbf{L}_{ori} + \mathbf{\lambda}_{rot} \mathbf{L}_{rot} + \mathbf{\lambda}_{fk} \mathbf{L}_{fk}
\end{equation}
We set the weights $\mathbf{\lambda}_{ori}$, $\mathbf{\lambda}_{rot}$, and $\mathbf{\lambda}_{fk}$ to 0.05, 1, and 1, respectively.
For fast training, we do not include our IK Module into the training stage.

\section{Experiments}
\subsection{Data Preparation and Network Training}
We use the subsets CMU~\cite{cmu},  BMLrub~\cite{troje2002decomposing} and HDM05~\cite{cg-2007-2} in AMASS~\cite{AMASS:ICCV:2019} dataset for training and testing. The AMASS dataset is a large human motion database that unifies different existing optical marker-based MoCap datasets by converting them into realistic 3D human meshes represented by SMPL~\cite{loper2015smpl} model parameters. We split the three datasets into random training and test sets with 90\% and 10\% of the data, respectively. For use on VR devices, we unified the frame rate to 60 Hz.

To optimize the parameters of \methodnospace, we adopt the Adam solver~\cite{kingma2015adam} with batch size 256. We set the chunk size of input as 40 frames. The learning rate starts from $1\times10^{-4}$ and decays by a factor of 0.5 every $2\times10^4$ iterations. We train our model with PyTorch on one NVIDIA GeForce GTX 3090 GPU. It takes about two hours to train \methodnospace.

\subsection{Evaluation Results}
We use MPJRE (Mean Per Joint Rotation Error [\degree]), MPJPE (Mean Per Joint Position Error [cm]), and MPJVE (Mean Per Joint Velocity Error [cm/s]) as our evaluation metrics. We compare our proposed \method with Final IK~\cite{finalik}, CoolMoves~\cite{ahuja2021coolmoves}, LoBSTr~\cite{yang2021lobstr}, and VAE-HMD~\cite{dittadi2021full}, which are state-of-the-art methods working on the problems of avatar pose estimation from sparse inputs.

\begin{table*}[tb]

    \centering
    
    \setlength{\tabcolsep}{5.0pt}
    \caption{Comparisons of MPJRE [\degree], MPJPE [cm], and MPJVE  [cm/s] to State-of-the-Arts on AMASS dataset. For each metric, the best result is highlighted in \textbf{boldface}.}
    \label{tab:sota}

    \begin{tabular}{@{}l rrr|rrr@{}}
        \toprule
                & \multicolumn{3}{c|}{Four Inputs}              & \multicolumn{3}{c}{Three Inputs}             \\

        Methods          & MPJRE & MPJPE & MPJVE & MPJRE & MPJPE & MPJVE\\
        \midrule
Final IK	            &	12.39	&	9.54	&	36.73	&	16.77	&	18.09	&	59.24\\
CoolMoves   	        &	4.58	&	5.55	&	65.28	&	5.20	&	7.83	&	100.54\\
LoBSTr      	        &	8.09	&	5.56	&	30.12	&	10.69    &	9.02	&	44.97\\
VAE-HMD     	        &	3.12	&	3.51	&	28.23	&	4.11	&	6.83	&	37.99\\
AvatarPoser (Ours)     &	\textbf{2.59}	&	\textbf{2.61}	&	\textbf{22.16}	&	\textbf{3.21}	&	\textbf{4.18}	&	\textbf{29.40}\\
        \bottomrule
    \end{tabular}
\end{table*}

\begin{figure}[tb]
    \centering
\begin{overpic}[width=\linewidth]{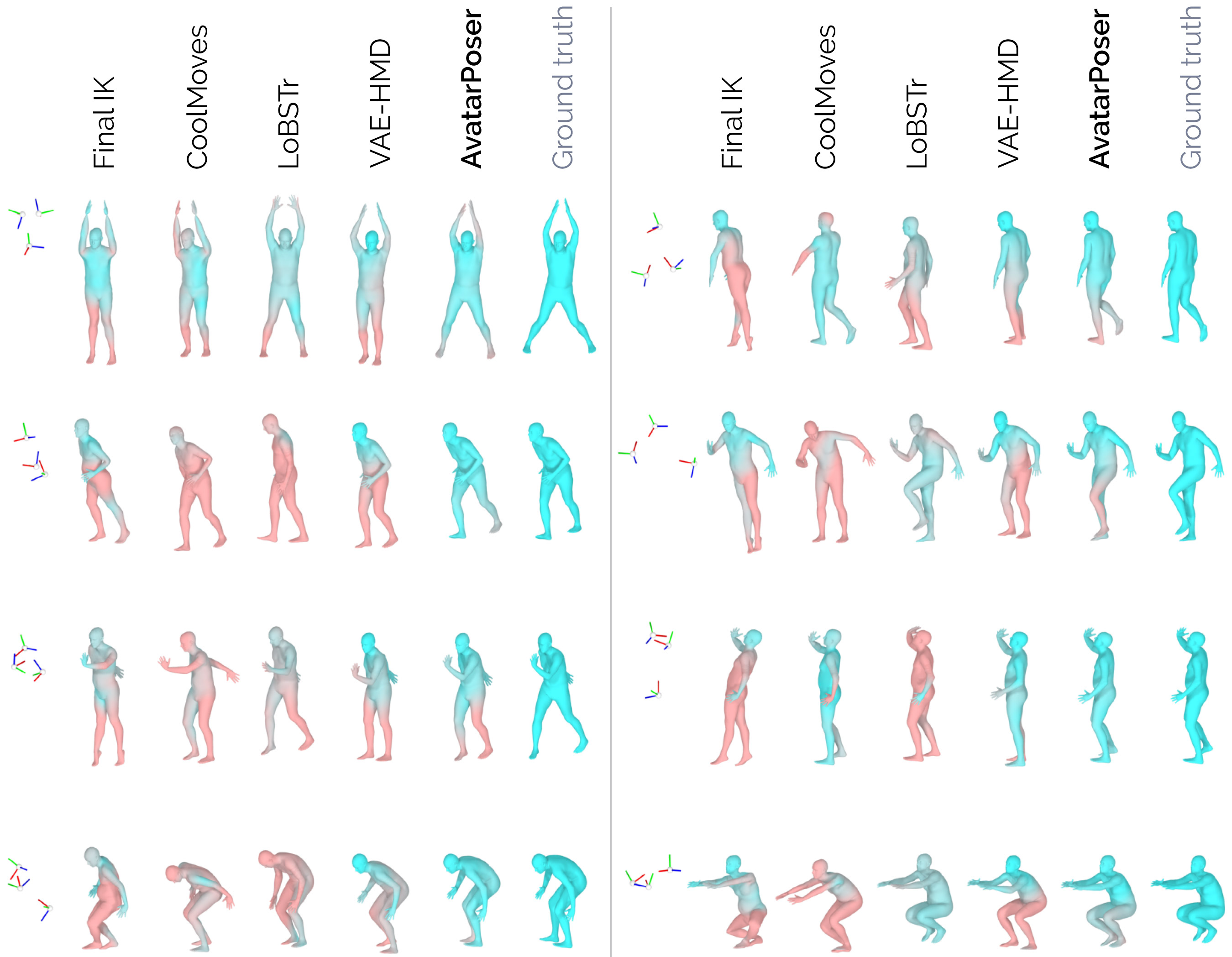}
\end{overpic}
    \caption{%
    Visual comparisons of different methods based on given sparse inputs for various motions. Avatars are color-coded to show errors in red.%
    }
    \label{fig:my_label}
\end{figure}

\begin{figure}[h!]
    \centering
    \subfigure[Final IK]{
    \includegraphics[width=0.310\linewidth, clip, trim=0.5cm 0.4cm 0.4cm 0.5cm]{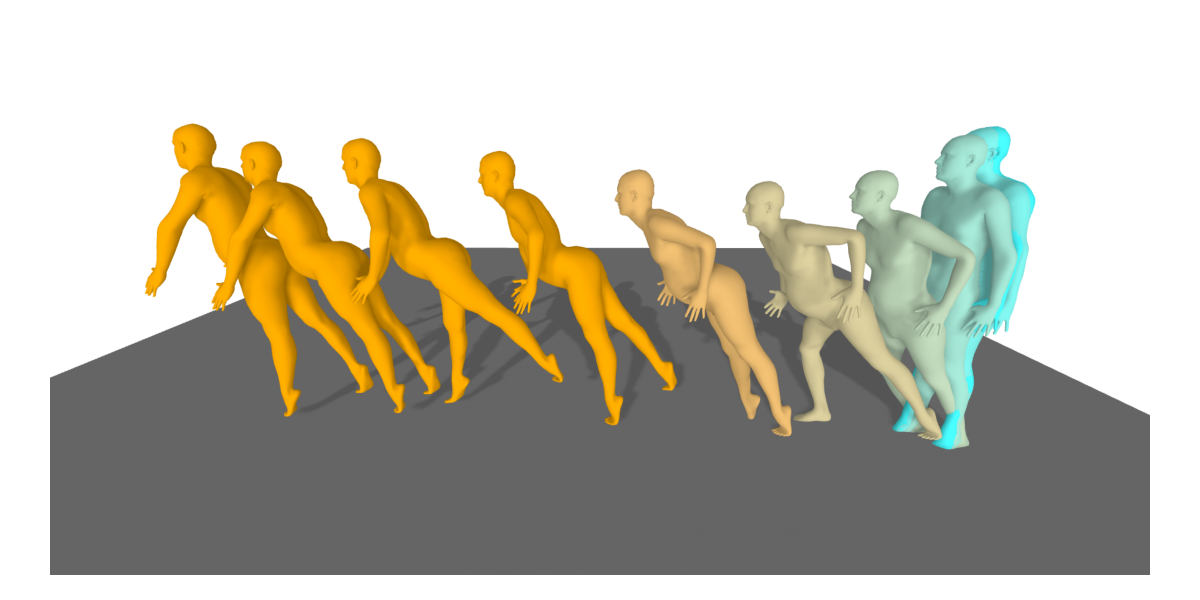}
    }
    \subfigure[CoolMoves]{
    \includegraphics[width=0.310\linewidth, clip, trim=0.5cm 0.4cm 0.4cm 0.5cm]{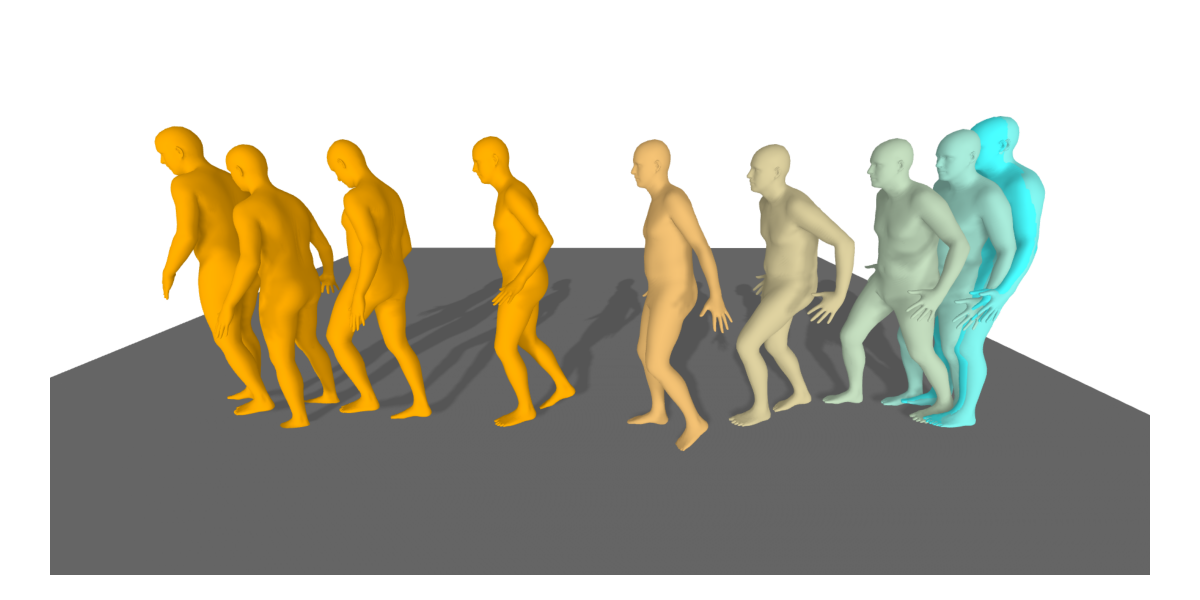}
    }
    \subfigure[LoBSTr]{
    \includegraphics[width=0.310\linewidth, clip, trim=0.5cm 0.4cm 0.4cm 0.5cm]{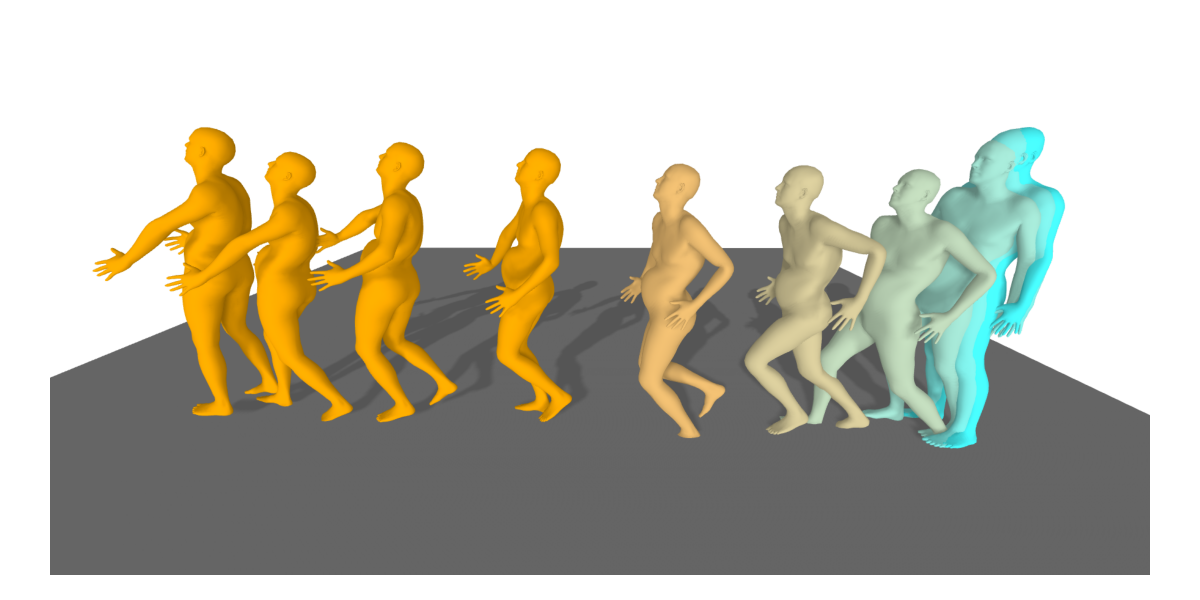}
    }\hfill
    \subfigure[VAE-HMD]{    
    \includegraphics[width=0.310\linewidth, clip, trim=0.5cm 0.4cm 0.4cm 0.5cm]{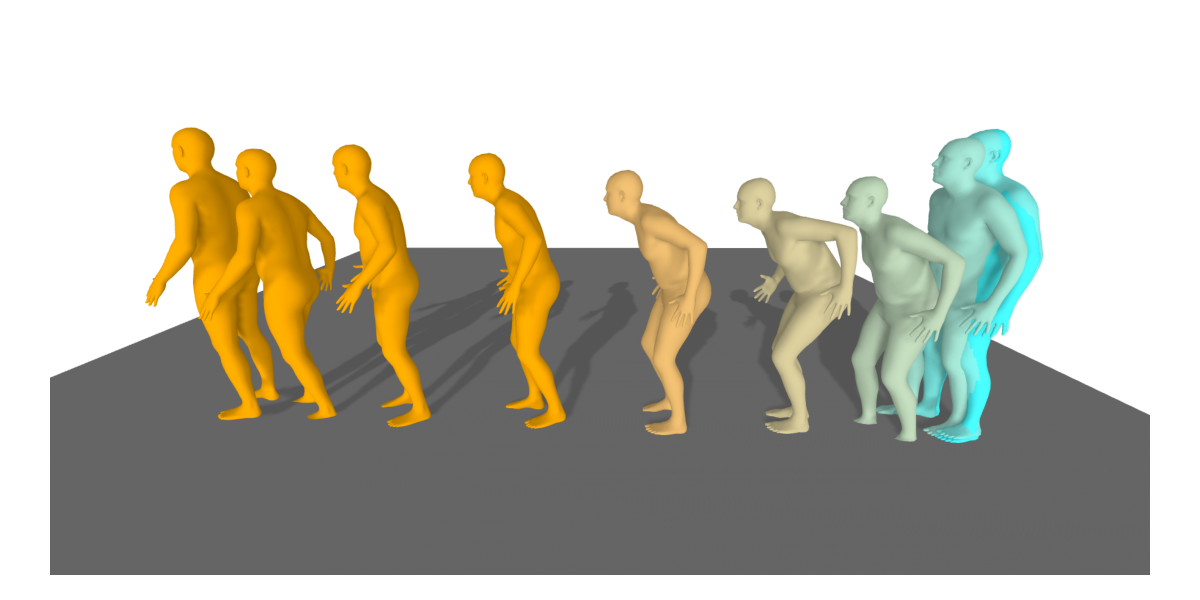}
    }
    \subfigure[AvatarPoser (Ours)]{
    \includegraphics[width=0.310\linewidth, clip, trim=0.5cm 0.4cm 0.4cm 0.5cm]{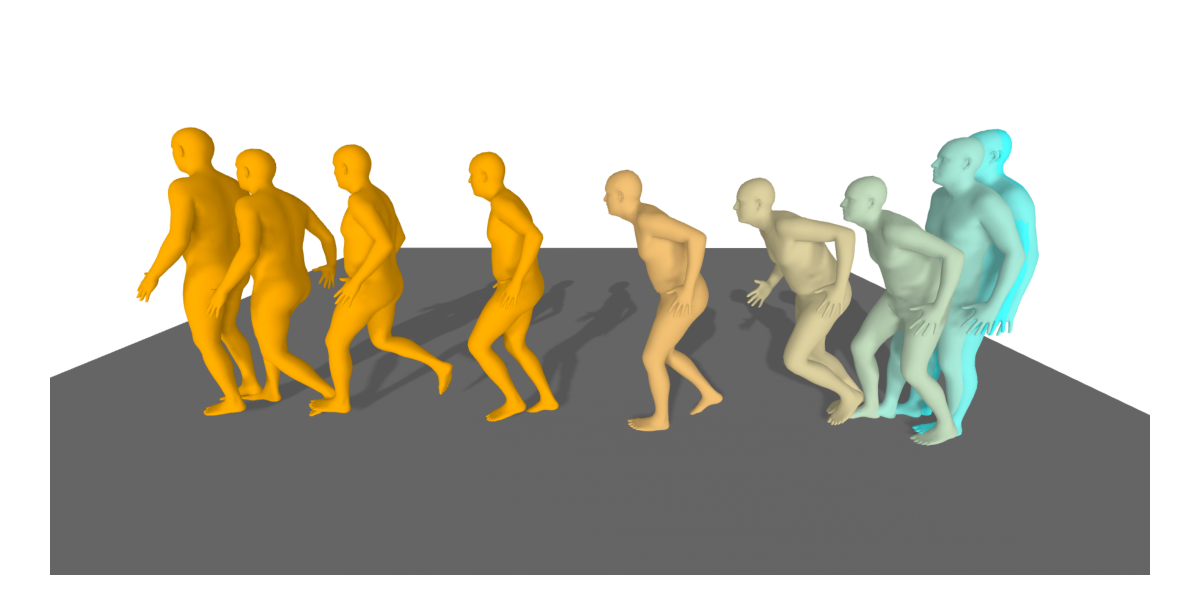}
    }
    \subfigure[Ground Truth]{
    \includegraphics[width=0.310\linewidth, clip, trim=0.5cm 0.4cm 0.4cm 0.5cm]{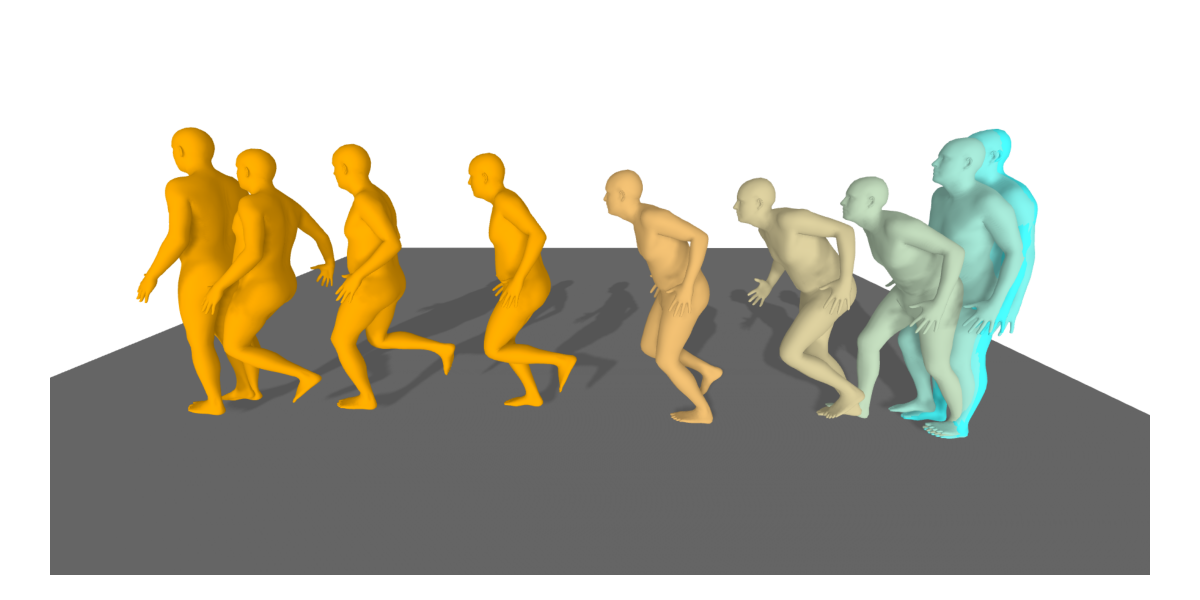}    
    }
    \caption{Visual results of our proposed method \method compared to SOTA alternatives for a running motion. The change of color denotes different timestamp.
    }
    \label{fig:sota}    
\end{figure}    
Since these state-of-the-art methods do not provide public source codes, we directly run Final IK in Unity~\cite{finalik} and reproduce other methods to the best of our knowledge. For a fair comparison, we train all the methods on the same training and testing data. It should be noted that the original CoolMoves is a position-based method, we adapt it to rotation-based method for a fair comparison with other methods. We make all the methods work with both three (headset, controllers) and four inputs (headset, controllers, pelvis tracker). When only three inputs are provided, for input representation we do not use the pose of pelvis as a reference frame, and for the output we calculate the global orientation and translation of human body at pelvis through the kinematic chains from the given global pose of head.

The numerical results for the considered metrics ({\small MRJRE}, {\small MPJPE}, and {\small MPJVE}) for both four and three inputs are reported in Table~\ref{tab:sota}. It can be seen that our proposed \method achieves the best results on all three metrics and outperforms all other methods. VAE-HMD achieves the second best performance on {\small MPJPE}, which is followed by CoolMoves (KNN). Final IK gives the worst result on {\small MPJPE} and {\small MPJRE} because it optimizes the pose of the end-effectors without considering the smoothness of other body joints. As a result, the performance of LoBSTr, which uses Final IK for upper body pose estimation, is also low. We believe this shows the value in data-driven methods to learn motion from existing mocap datasets. However, it does not mean that traditional optimization methods are not useful. In our ablation studies, we show how inverse kinematics when combined with deep learning can improve the accuracy of hand positions.

To further evaluate the generalization ability of our proposed method, we perform a 3-fold cross-dataset evaluation among different methods. To do so, we train on two subsets and test on the other subset in a round robin fashion. Table~\ref{tab:crossdataset} shows the experimental results of different methods tested on CMU, BMLrub, and HDM05 datasets. We achieve the best results over almost all evaluation metrics in all three datasets. Although Final IK performs slightly better than \method in terms of MPJVE in CMU, which can only means the motions are a little bit smoother. However, the rotation error MPJRE and the position error MPJPE of Final IK, which represent the accuracy of predictions, are much larger than our method.

\begin{table*}[]

    \centering
    
    \setlength{\tabcolsep}{8.0pt}
    \caption{Results of cross-dataset evaluation between different methods. The input signals are from only three devices, i.e., one headset and two controllers. The best results for each dataset and each evaluation metrics are highlighted in \textbf{boldface}.}
    \label{tab:crossdataset}

    \begin{tabular}{@{}ll rrr@{}}
        \toprule
    Dataset    &Methods          & MPJRE & MPJPE & MPJVE \\
        \midrule
\multirow{ 5}{*}{CMU}&Final IK	            &	17.80	&	18.82	&	56.83\\
&CoolMoves   	        &	9.20	&	18.77	&	139.17\\
&LoBSTr      	        &	12.51	&	12.96	&	49.94\\
&VAE-HMD     	        &	6.53	&	13.04	&	51.69\\
&AvatarPoser (Ours)     &	\textbf{5.93}	&	\textbf{8.37}	&	\textbf{35.76}\\
        \midrule
\multirow{ 5}{*}{BMLrub}&Final IK	            &	15.93	&	17.58	&	60.64\\
&CoolMoves   	        &	7.93	&	13.30	& 134.77\\
&LoBSTr      	        &	10.79	&	11.00	&	60.74\\
&VAE-HMD     	        &	5.34	&	9.69	&	51.80\\
&AvatarPoser (Ours)     &	\textbf{4.92}	&	\textbf{7.04}	&	\textbf{43.70}\\
        \midrule
\multirow{ 5}{*}{HDM05}&Final IK	            &	18.64	&	18.43	&	62.39\\
&CoolMoves   	        &	9.47	&	17.90	&	140.61\\
&LoBSTr      	        &	13.17	&	11.94	&	48.26\\
&VAE-HMD     	        &	6.45	&	10.21	&	40.07\\
&AvatarPoser (Ours)     &	\textbf{6.39}	&	\textbf{8.05}	&	\textbf{30.85}\\
        \bottomrule
    \end{tabular}
\end{table*}

\subsection{Ablation Studies}

We perform an ablation study on the different submodules of our method and provide results in Table~\ref{tab:ablation}. The experiments are conducted on the same test set as HDM05  in Table~\ref{tab:crossdataset}. We use {\small MPJRE} [\degree], {\small MPJPE} [cm] as our evaluation metrics in the ablation studies to show the need for each component. In addition to the position error across the full-body joints, we specifically calculate the mean error on hands to show how the IK module helps improve the hand positions. \mbox{}\\

\begin{table*}[t]

    \centering
    
    \setlength{\tabcolsep}{8.0pt}
    \caption{Ablation studies. Best results are highlighted in \textbf{bold} for each metric.}
    \label{tab:ablation}

    \begin{tabular}{@{}l rrr@{}}
        \toprule

        Configurations          & MPJRE & MPJPE-Full Body & MPJPE-Hand \\
        \midrule
Default	                        &	6.39	&	\textbf{8.05}	&	\textbf{1.86}	\\
No Stabilizer   	            &	6.39	&	9.29	&	2.15	\\
Predict Pelvis Position         &	6.42	&	8.82	&	2.11	\\
No FK Module     	            &	\textbf{6.24}	&	8.41	&	2.04	\\
No IK Module                    &	6.41	&	8.07	&	3.17    \\
        \bottomrule
    \end{tabular}
\end{table*}

\noindent\textbf{No Stabilizer.} We remove the Stabilizer module, which predicts the global orientation, and calculate the global orientation through the body kinematic chain directly from the given orientation of the head. Table~\ref{tab:ablation} shows that the {\small MPJPE} drops without Stabilizer. This is because the rotation of the head is relatively independent to the rest of the body. Therefore, the global orientation is highly sensitive to random rotations of the head. Learning the global orientation from richer information via the network is a superior way to solve the problem. \mbox{}\\

\noindent\textbf{Predict Pelvis Position.} In our final model, we calculate the global translation of the human body, which is located at the pelvis, from the input head position through the kinematic chain. We also try directly regressing to the global translation within the network, but the result is worse than computing via the kinematic chain according to our evaluation results.  \mbox{}\\

\noindent\textbf{No FK Module. } We also remove the FK Module, which means the network is only trained to minimize the rotation angles without considering the positions of joints after forward kinematics calculation. When we remove the FK module, the {\small MPJPE} increases and the MPJRE decreases. This is intuitive as we only optimize the joint rotations without the IK module.  While rotation-based methods provide robust results without the need to reproject onto skeleton constraints to avoid bone stretching and invalid configurations, they are prone to error accumulation along the kinematic chain.  \mbox{}\\

\noindent\textbf{No IK Module.} We remove IK Module and only provide the results directly predicted by our neural network. Removing the IK module has little effect on the average position error of full-body joints. However, the average position error of the hands increases by almost 41\%. 

\subsection{Running Time Analysis}

\begin{figure}
    \centering
    \includegraphics[trim=0cm 0.1cm 0cm 1.2cm,clip=true,width=0.6\textwidth]{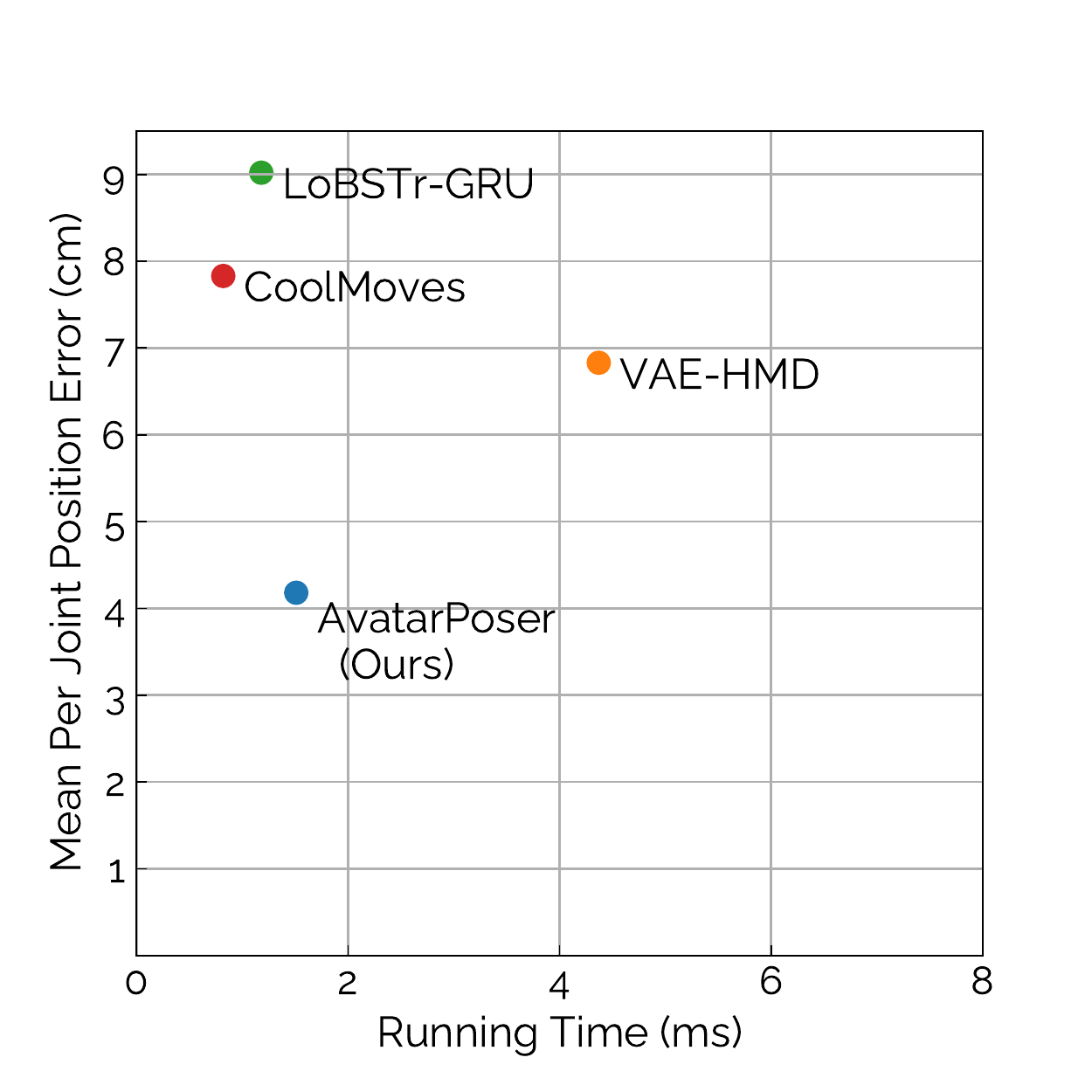}
    \caption{Comparisons of {\textbf network} inference time across several methods. Due to the powerful and efficient Transformer encoder, our method achieves the smallest average position error while providing fast inference.}
    \label{fig:runtime}
\end{figure}

We evaluated the run-time inference performance of our network \method and compared it to the inference of VAE-HMD~\cite{dittadi2021full}, LoBSTr~\cite{yang2021lobstr}, CoolMoves~\cite{ahuja2021coolmoves} as shown in Fig.~\ref{fig:runtime}.
Note that we did not include Final IK~\cite{finalik}, because its integration into Unity makes accurate measurements difficult.
To conduct our comparison, we modified LoBSTr to directly predict full-body motion via the GRU (denoted as LoBSTr-GRU) instead of combining Final IK and the GRU together.
We measured the run time per frame (in milliseconds) on the evaluated test set on one NVIDIA 3090 GPU.
For a fair comparison, we only calculated the network inference time of \method here.
Our \method achieves a good trade-off between performance and inference speed.

Our method also requires executing an IK algorithm after the network forward pass. Each  iteration costs approximately 6\,ms, so we set the number of iterations to 5 to keep a balance between inference speed and the accuracy of the final hand position.
Note that the speed could be accelerated by adopting a more standard non-linear optimization.

\subsection{Test on a Commercial VR System}

\begin{figure}
    \centering
    \includegraphics[width=\linewidth]{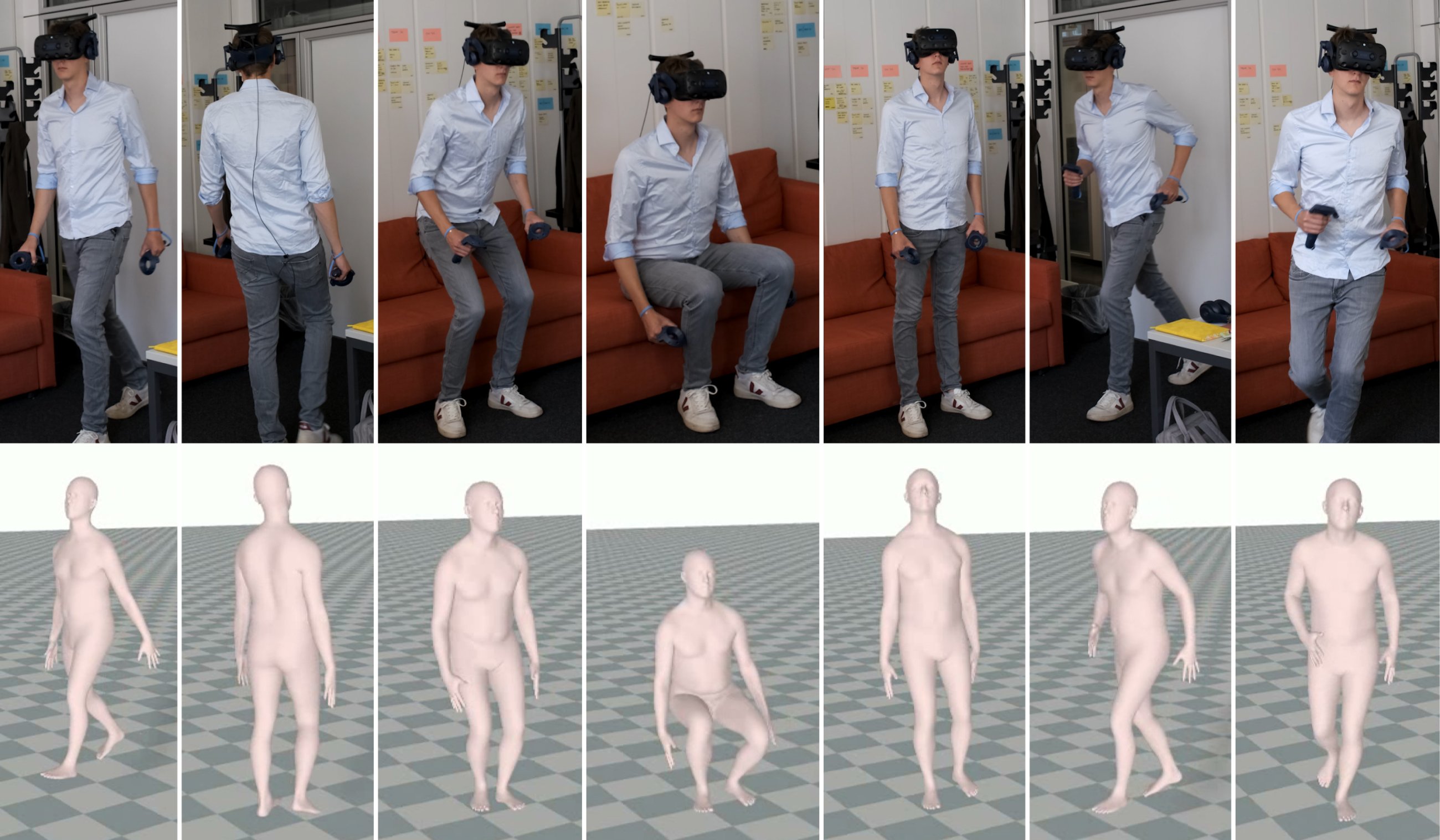}
    \caption{%
        We tested our method on recorded motion data from a VIVE Pro headset and two VIVE controllers.
        Columns show the user's pose (top) and our prediction (bottom).%
    }
    \label{fig:real_vr}
\end{figure}

To qualitatively assess the robustness of our method, we executed our algorithm on live recordings from an actual VR system.
We used an HTC VIVE HMD as well as two controllers, each providing real-time input with six degrees of freedom (rotation and translation).
Fig.~\ref{fig:real_vr} shows a few examples of our method's output based on sparse inputs.

\section{Conclusions}

We presented our novel Transformer-based method \method to estimate realistic human poses from just the motion signals of a Mixed Reality headset and the user's hands or hand-held controllers.
By decoupling the global motion information from learned pose features and using it to guide pose estimation, we achieve robust estimation results in the absence of pelvis signals.
By combing learning-based methods with traditional model-based optimization, we keep a balance between full-body style realism and accurate hand control.
Our extensive experiments on the AMASS dataset demonstrated that \method surpasses the performance of state-of-the-art methods and, thus, provides a useful learning-based IK solution for practical VR/AR applications.\mbox{}\\

\noindent\textbf{Acknowledgments:}
We thank Christian Knieling for his early explorations of learning-based methods for pose estimation with us at ETH Zürich.
We thank Zhi Li, Xianghui Xie, and Dengxin Dai from Max Planck Institute for Informatics for their helpful discussions.
We also thank Olga Sorkine-Hornung and Alexander Sorkine-Hornung for early discussions.

\clearpage
%
%
\bibliographystyle{splncs04}
\bibliography{literature}
\end{document}